\documentclass[lettersize,journal]{IEEEtran}
\usepackage{amsmath,amsfonts}
\usepackage{algorithmic}
\usepackage{algorithm}
\usepackage{array}
\usepackage{textcomp}
\usepackage{stfloats}
\usepackage{url}
\usepackage{verbatim}
\usepackage{graphicx}
\usepackage{cite}
\usepackage{booktabs}
\usepackage{multirow}
\usepackage{multicol}
\usepackage{hyperref}
\usepackage{placeins}
\usepackage{subfigure}
\usepackage{wrapfig}
\usepackage[table]{xcolor}
\begin{document}

\title{{MacroNav: Multi-Task Context Representation Learning Enables Efficient Navigation in Unknown Environments}}

\author{
Kuankuan Sima, Longbin Tang, Zhenyu Yang, Haozhe Ma, and Lin Zhao%
\thanks{Manuscript received November 6, 2025; accepted April 7, 2026. This paper was recommended for publication by Editor Aniket Bera upon evaluation of the Associate Editor and Reviewers' comments. \textit{(Corresponding author: Lin Zhao.)}}%
\thanks{Kuankuan Sima, Longbin Tang, and Lin Zhao are with the Department of Electrical and Computer Engineering; Zhenyu Yang is with the Department of Mechanical Engineering; and Haozhe Ma is with the School of Computing, National University of Singapore, Singapore 119077, Singapore (e-mail: \tt \footnotesize {kuankuan\_sima@u.nus.edu; elezhli@nus.edu.sg).}}%
\thanks{The code is available at: \href{https://github.com/SimonKennethRobo/MacroNav}{https://github.com/SimonKennethRobo/MacroNav}.}
\thanks{Digital Object Identifier (DOI): see top of this page.}
}

\markboth{IEEE ROBOTICS AND AUTOMATION LETTERS. PREPRINT VERSION. ACCEPTED Apr, 2026}
{Shell \MakeLowercase{\textit{et al.}}: A Sample Article Using IEEEtran.cls for IEEE Journals}


\maketitle

\begin{abstract}
\textcolor{black}{
Autonomous navigation in unknown environments requires multi-scale spatial understanding that captures geometric details, topological connectivity, and global structure to support high-level decision making under partial observability. Existing approaches struggle to efficiently capture such multi-scale spatial understanding while maintaining low computational cost for real-time navigation.
}
We present \textit{MacroNav}, a learning-based navigation framework featuring two key components: (1) a lightweight context encoder trained via multi-task self-supervised learning to capture multi-scale, navigation-centric spatial representations; and (2) a reinforcement learning policy that seamlessly integrates these representations with graph-based reasoning for efficient action selection. Extensive experiments demonstrate the context encoder’s effective and robust environmental understanding. Real-world deployments further validate \textit{MacroNav}’s effectiveness, yielding significant gains over state-of-the-art navigation methods in both Success Rate (SR) and Success weighted by Path Length (SPL), with superior computational efficiency.
\end{abstract}


\section{Introduction}
\IEEEPARstart{A}utonomous robot navigation in unknown environments remains a fundamental challenge in robotics, requiring real-time decision-making under partial observability while ensuring collision-free paths to specified goals. Classical methods such as A* and RRT* excel in pre-mapped environments but struggle with dynamic exploration scenarios where environmental structure must be incrementally discovered.

Based on advancements in Simultaneous Localization and Mapping
(SLAM) techniques, robots can obtain accurate real-time state estimations while building maps, enabling autonomous navigation in unknown environments. Previous works have explored frontier-based methods~\cite{dasilvalubancoNovelFrontierBasedExploration2020} and utility-cost assignments~\cite{holzEvaluatingEfficiencyFrontierbased2010}. Other approaches focus on continuous environment representation updates, such as FAR Planner~\cite{yang_far_2022} which builds visibility graphs. \textcolor{black}{However, these methods heavily rely on hand-crafted rules, including manually designed frontier selection heuristics~\cite{dasilvalubancoNovelFrontierBasedExploration2020}, predefined utility-cost trade-off functions~\cite{holzEvaluatingEfficiencyFrontierbased2010}, and fixed graph construction and pruning strategies~\cite{yang_far_2022}. Such rules encode rigid, scenario-specific assumptions rather than capturing the semantic and structural complexity of real-world environments, resulting in insufficient environmental understanding and suboptimal navigation efficiency}.

\begin{figure}[t]
\centering
\includegraphics[width=0.45\textwidth]{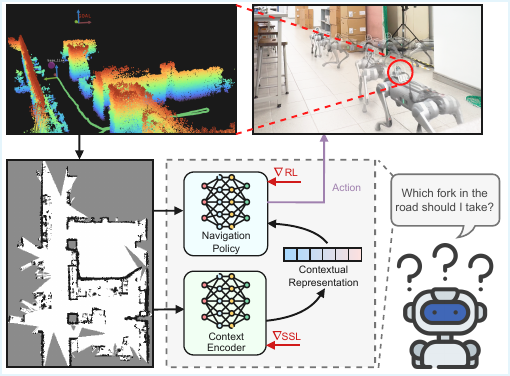}
\caption{Effective contextual representation facilitates cognition and reasoning for navigation. We propose a context encoder trained via multi-task self-supervised learning operating at different spatial scales. An RL-based navigation policy seamlessly integrates the learned contextual representations to select the optimal waypoint.}
\label{fig:fig1} \vspace{-0.5cm}
\end{figure}

\textcolor{black}{
The advent of deep reinforcement learning (DRL) has transformed navigation research by enabling learned planning without hand-crafted rules. While end-to-end RL algorithms such as DD-PPO~\cite{wijmans2019dd} and NavRL~\cite{xu2025navrl} demonstrate success in constrained environments, they exhibit limitations in unseen complex environments due to lack of explicit {context} modeling. Here, \textit{context} refers to the agent's global understanding of the observed environment, capturing spatial dependencies across different regions to enable non-myopic decision-making beyond local greedy strategies. Recently, methods such as HD Planner~\cite{liang2024hdplanner} and~\cite{caoDeepReinforcementLearningBased2024} have shown that explicitly modeling such context using topological graphs with attention mechanisms substantially outperforms classical methods, as attention layers capture long-range dependencies between regions and allow the agent to reason about potential structures in unexplored areas. However, their node-based topological representations abstract continuous spaces into discrete nodes and edges, discarding fine-grained geometric information such as corridor structures, passage configurations, and spatial layouts between regions. This limits the agent's ability to fully exploit spatial context for reasoning, leading to less informed decision-making in geometrically complex environments.
}

\textcolor{black}{
Alternative approaches explore dense representations using CNN~\cite{he2024alpha} and ViT~\cite{liu2023vit} backbones. Direct application of pre-trained visual models suffers from domain shift between natural images and navigation maps~\cite{zhou2022domain}. To address this, some methods explore self-supervised methods for visual representation, exemplified by OVRL~\cite{yadav2023offline}. Despite showing promise, these adopt generic self-supervised objectives (e.g., DINO~\cite{oquab2023dinov2}) without considering navigation-specific characteristics. These limitations highlight the need for a contextual representation that is both dense—preserving geometric structure and spatial continuity—and navigation-aware, capturing task-relevant environmental properties.
}

Our key insight is that effective navigation requires three complementary capabilities: (1) global environmental structure understanding, (2) local geometric continuity recognition, and (3) occlusion-robust representation. \textcolor{black}{While not exhaustive, these are widely recognized as core requirements for navigation in unknown environments~\cite{wijayathunga2023challenges}, and each targets a distinct, non-substitutable aspect of the problem. Existing single-task self-supervised learning methods fail to simultaneously capture these requirements.
}

We propose \textbf{M}ulti-t\textbf{A}sk \textbf{C}ontext \textbf{R}epresentati\textbf{O}n learning for \textbf{Nav}igation (MacroNav), which synergistically combines three SSL tasks at different spatial scales: (1) Stochastic Path Masking (SPM): predicting spatially coherent masked regions via random walk trajectories to capture long-range dependencies and global layouts; (2) Field-of-View (FOV) Prediction: inferring peripheral regions for local geometric continuity recognition; (3) Masked Autoencoding (MAE): reconstructing heavily masked areas for occlusion robustness. We further propose an RL-based navigation policy that efficiently fuses learned contextual representations with topological graph reasoning through cross-attention. As illustrated in Figure~\ref{fig:fig1}, this design substantially improves navigation efficiency in complex unknown environments. Our main contributions are:
\begin{itemize}
\item A multi-task self-supervised representation learning method that addresses three complementary requirements for navigation: global structure understanding, local geometric recognition, and occlusion robustness.

\item A context-aware RL navigation policy that integrates learned multi-scale contextual representations with explicit topological graph modeling through hierarchical cross-attention, achieving superior reasoning capability compared to node-centric and end-to-end approaches.

\item Comprehensive experimental validation demonstrating that our representations have better spatial understanding capabilities than common pre-trained visual models, and real-world experiments showing significant improvements over state-of-the-art (SotA) navigation methods.
\end{itemize}

\section{Related Works}

\textbf{Classical Navigation.}
Traditional methods such as A* and RRT*~\cite{zhou2022review} excel in pre-mapped environments but struggle with real-time exploration. Frontier-based strategies~\cite{dasilvalubancoNovelFrontierBasedExploration2020} systematically identify unmapped regions, though target prioritization relies on manual heuristics~\cite{holzEvaluatingEfficiencyFrontierbased2010}. FAR Planner~\cite{yang_far_2022} maintains continuous representation via visibility graphs but incurs significant computational overhead in complex scenarios. These methods generally exhibit limited adaptability due to their reliance on hand-crafted rules.

{\color{black}
\textbf{Learning-Based Navigation.}
Deep reinforcement learning enables data-driven navigation without hand-crafted heuristics. End-to-end approaches such as DD-PPO~\cite{wijmans2019dd}, NavRL~\cite{xu2025navrl}, and Fast-LiDARNet~\cite{liu2021efficient} achieve strong results in controlled settings but generalize poorly to unseen environments due to implicit representations learned jointly with control. For example, NavRL lacks explicit global planning in complex topologies, while Fast-LiDARNet depends on supervised demonstrations and targets lane-following tasks.

Graph-based methods~\cite{liang2024hdplanner,caoDeepReinforcementLearningBased2024} introduce topological reasoning, but their discrete node features omit fine-grained geometric details needed in cluttered spaces. Dense representations using CNN~\cite{he2024alpha} and ViT~\cite{liu2023vit} backbones suffer from domain mismatch between natural images and occupancy maps~\cite{zhou2022domain}. Self-supervised approaches such as OVRL~\cite{yadav2023offline} rely on generic objectives like DINO~\cite{oquab2023dinov2} without navigation-specific design, while NavRep~\cite{dugas2021navrep} focus on dynamic prediction rather than structural understanding.

}

\section{Method}
\label{sec:method}

\begin{figure*}[t]
\centering
\includegraphics[width=0.85\textwidth]{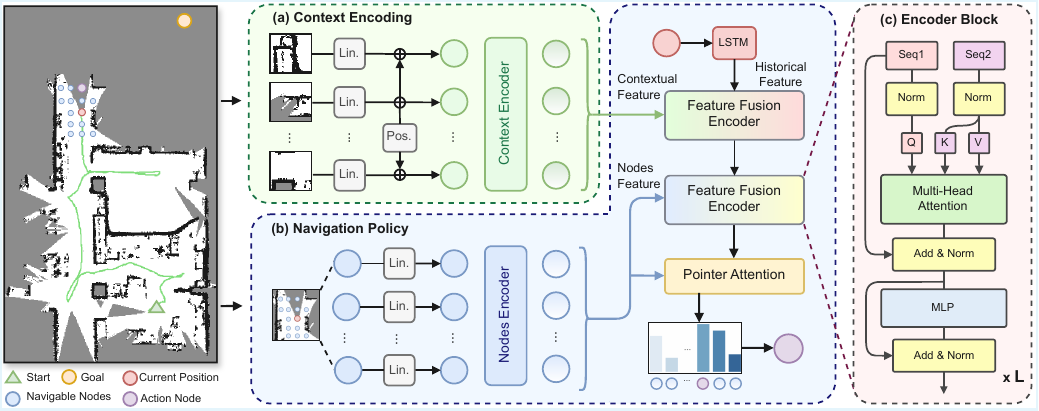}
\caption{Overall architecture of \textit{MacroNav}. (a) The context map is tokenized and processed by the pre-trained context encoder to extract spatial representations. (b) Navigable nodes are encoded and fused with contextual features through cross-attention, followed by pointer attention to select the action node. (c) All encoders are based on the multi-layer multi-head attention mechanisms.}
\label{fig:overall}
\end{figure*}

We present \textit{MacroNav}, a context-aware navigation framework for autonomous navigation in unknown environments. As illustrated in Figure~\ref{fig:overall}, our approach comprises: (1) a context encoder extracting multi-scale spatial representations from occupancy maps, (2) a navigation policy selecting optimal waypoints by fusing contextual representations with graph-based reasoning through cross-attention, and (3) a transformer-based~\cite{vaswaniAttentionAllYou2017} architecture enabling effective feature integration.

Our key innovation is the synergistic combination of navigation-specific self-supervised context representation learning with context-aware policy learning. We first pre-train the context encoder using our multi-task self-supervised learning method (Section~\ref{sec:ssl}), then integrate it into an RL-based navigation policy (Section~\ref{sec:policy}).

\subsection{Context Encoder Architecture}
\label{sec:encoder}

The context encoder $f_\theta$ processes the context map $\mathcal{M}_t \in \mathbb{R}^{H \times W}$ to extract spatial representations, where $H$ and $W$ denote the spatial dimensions. We partition the context map into a sequence of non-overlapping patches, resulting in flattened 2D patch tokens $\mathcal{M}_t^p \in \mathbb{R}^{N\times P^2}$, $(P,P)$ represents the patch resolution, and $N=HW/P^2$ is the total number of patches.

Each patch is linearly projected to a $d$-dimensional embedding and augmented with learnable positional encodings before being fed into the encoder:
\begin{equation}
\mathbf{z}_0 = \text{Proj}(\mathcal{M}_t^p) + \mathbf{E}_{\text{pos}}.
\end{equation}

The encoder consists of $L$ stacked transformer layers. Each layer $\ell \in \{1, \ldots, L\}$ applies multi-head self-attention (MHSA) followed by a feed-forward network (FFN), with residual connections and layer normalization (LN):
\begin{equation}
\begin{aligned}
\mathbf{z}'_\ell &= \text{LN}(\mathbf{z}_{\ell-1} + \text{MHSA}(\mathbf{z}_{\ell-1})), \\
\mathbf{z}_\ell &= \text{LN}(\mathbf{z}'_\ell + \text{FFN}(\mathbf{z}'_\ell)).
\end{aligned}
\end{equation}
MHSA follows the standard formulation in~\cite{vaswaniAttentionAllYou2017}; cross-attention used later in the policy shares the same mechanism but derives queries from a different sequence than keys and values.

\subsection{Self-Supervised Context Representation Learning}
\label{sec:ssl}

Effective navigation in unknown environments demands context representations that capture three critical properties: (1) \textit{global structure understanding} for long-range spatial dependencies, (2) \textit{local geometric continuity recognition} for fine-grained navigational cues, and (3) \textit{occlusion robustness} to handle severe partial observability in real-world scenarios.

\begin{figure}[t]
\centering
\includegraphics[width=0.4\textwidth]{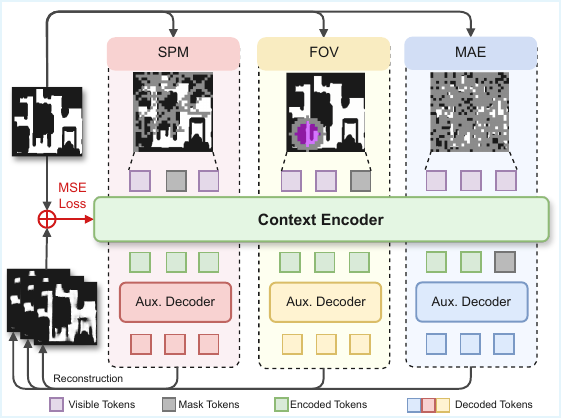}
\caption{Overview of the multi-task self-supervised learning method comprising three complementary tasks operating at different spatial scales.}
\label{fig:ssl}\vspace{-0.3cm}
\end{figure}

Inspired by recent advances in masked visual representation learning~\cite{mae}, we propose a multi-task self-supervised learning method tailored for navigation context encoding. As illustrated in Fig.~\ref{fig:ssl}, our pre-training pipeline comprises three complementary tasks: (1) \textit{Stochastic Path Masking (SPM)} for global layout understanding, (2) \textit{Field-of-View (FOV) Prediction} for local geometric continuity recognition, and (3) \textit{Masked Autoencoding (MAE)} for robustness against partial observability.

\subsubsection{Stochastic Path Masking}

To learn global layout understanding, we propose Stochastic Path Masking (SPM), a novel masking strategy that generates spatially continuous yet stochastically irregular masked regions through random walks on the patch grid. Unlike random or block-wise masking in traditional vision tasks, SPM aligns with navigation scenarios exhibiting spatial continuity in agent movement and environmental structure. We opt for random walks over path planning algorithms to ensure training efficiency.

Starting from a random patch, the walk proceeds for $T = \lfloor \rho \cdot N\rfloor$ steps, marking visited patches as masked. At each step, smoothness parameter $s \in [0,1]$ controls trajectory persistence. With probability $s$, the walk maintains its previous direction; otherwise, it selects a new direction from an 8-directional neighborhood. During training, we use random sampling for both $\rho$ and $s$ to enhance generalization. The random walk generates a sequence of masked patches. We denote $\mathcal{M}_{\text{SPM}}$ as the index set for the masked patches. For each masked patch, we introduce a learnable mask token $\mathbf{m}_{\text{SPM}}$ as placeholder. 

Visible patch embeddings $\mathcal{P}_{\text{vis}}$ and mask tokens are concatenated and jointly processed by the encoder:
\begin{equation}
\mathbf{z}_{\text{full}} = f_\theta(\text{Concat}(\mathcal{P}_{\text{vis}}, \mathbf{m}_{\text{SPM}}) + \mathbf{E}_{\text{pos}}),
\end{equation}
where $\mathbf{z}_{\text{full}} \in \mathbb{R}^{N \times d}$ represents the complete sequence of encoded representations for all $N$ patches. A lightweight decoder then projects these representations to reconstruct the original context map. Let $\hat{\mathcal{M}}_{i}$ denote the reconstructed patch and $\mathcal{M}_{t,i}$ the corresponding ground truth from $\mathcal{M}_t$. The training loss is computed by comparing reconstructed patches against their corresponding ground-truth ones:
\begin{equation}
\mathcal{L}_{\text{SPM}} = \frac{1}{|\mathcal{M}_{\text{SPM}}|} \sum_{i \in \mathcal{M}_{\text{SPM}}} \|\hat{\mathcal{M}}_{i} - \mathcal{M}_{t,i}\|_2^2.
\end{equation}

\subsubsection{Field-of-View Prediction}

While SPM captures global structure, navigation also requires fine-grained understanding of local geometric patterns such as corridors, doorways, and passage widths. We introduce FOV prediction to explicitly learn local geometric continuity within confined spatial regions.

We randomly select a patch as the center of a circular core FOV region $\mathcal{S}_{\text{FOV}}$ with area ratio $\rho_{\text{FOV}}$ relative to the total number of patches $N$. The corresponding radius in patch units is computed as $r_{\text{FOV}} = \sqrt{\rho_{\text{FOV}} \cdot N / \pi}$. Similarly, an expanded circular region $\mathcal{S}_{\text{expand}}$ is defined with radius $r_{\text{expand}} = \sqrt{(\rho_{\text{FOV}} + \rho_{\text{expand}}) \cdot N / \pi}$. Patches within distance $r_{\text{FOV}}$ from the center remain visible, while patches in the expansion ring $\mathcal{S}_{\text{predict}} = \mathcal{S}_{\text{expand}} \setminus \mathcal{S}_{\text{FOV}}$ are masked and must be predicted. We denote $\mathcal{M}_{\text{FOV}}$ as the index set of masked patches in the expansion ring.

Similar to SPM, learnable mask tokens $\mathbf{m}_{\text{FOV}}$ are assigned to each masked patch in $\mathcal{M}_{\text{FOV}}$. Visible FOV patches $\mathcal{P}_{\text{FOV}}$ and mask tokens are fed to the encoder:
\begin{equation}
\mathbf{z}_{\text{full}} = f_\theta(\text{Concat}(\mathcal{P}_{\text{FOV}}, \mathbf{m}_{\text{FOV}}) + \mathbf{E}_{\text{pos}}).
\end{equation}
The decoder reconstructs masked patches in the expansion ring, producing $\hat{\mathcal{M}}_{i}$ for each $i \in \mathcal{M}_{\text{FOV}}$. The prediction loss is:
\begin{equation}
\mathcal{L}_{\text{FOV}} = \frac{1}{|\mathcal{M}_{\text{FOV}}|} \sum_{i \in \mathcal{M}_{\text{FOV}}} \|\hat{\mathcal{M}}_{i} - \mathcal{M}_{t,i}\|_2^2.
\end{equation}

\subsubsection{Masked Autoencoding}

To develop robustness against severe partial observability, we employ masked autoencoding with aggressive masking (75\% mask ratio). We randomly select patches to mask, forming the masked patch index set $\mathcal{M}_{\text{MAE}}$ where $|\mathcal{M}_{\text{MAE}}| = 0.75N$. Critically, only visible patches are processed by the encoder:
\begin{equation}
\mathbf{z}_{\text{vis}} = f_\theta(\text{Proj}(\mathcal{P}_{\text{vis}}) + \mathbf{E}_{\text{pos}}),
\end{equation}
where $\mathbf{z}_{\text{vis}}$ only contains representations for the visible patches. This asymmetric design forces the encoder to extract compact, information-dense representations crucial for decision-making under limited sensory information. 

For decoding, we interleave encoded visible tokens $\mathbf{z}_{\text{vis}}$ with learnable mask tokens $\mathbf{m}_{\text{MAE}}$ according to their original spatial positions, forming a full sequence $\mathbf{z}_{\text{full}} \in \mathbb{R}^{N \times d}$. This sequence is then processed by a lightweight decoder to produce reconstructions $\hat{\mathcal{M}}_{i}$ for all patches. The loss is computed only over the masked patches:
\begin{equation}
\mathcal{L}_{\text{MAE}} = \frac{1}{|\mathcal{M}_{\text{MAE}}|} \sum_{i \in \mathcal{M}_{\text{MAE}}} \|\hat{\mathcal{M}}_{i} - \mathcal{M}_{t,i}\|_2^2.
\end{equation}

\subsubsection{Multi-Task Joint Training}

For multi-task joint training, one task is sampled with equal probability at each step and only its loss is backpropagated, avoiding per-task weight tuning~\cite{crawshaw2020multi}. This yields multi-scale contextual representations combining global reasoning, local geometric recognition, and occlusion robustness for downstream policy learning.

\subsection{Context-Aware Navigation Policy}
\label{sec:policy}

We propose a reinforcement learning-based navigation policy that selects the optimal waypoint by explicitly constructing a local topological graph while integrating learned contextual representations through hierarchical cross-attention.

\subsubsection{State Representation} 
At each time step $t$, centered on the agent's current position $p_t$, we sample $K$ collision-free candidate waypoints uniformly within a local neighborhood $r_{local}$ to form a node set $V_t = \{v_1, v_2, \ldots, v_K\}$. These nodes are connected via $k$-nearest neighbors (kNN) to construct an undirected topological graph $G_t = (V_t, E_t)$. Each navigation node $v_i$ is characterized by a feature tuple $v_i = (\vec{n}_i, u_i, \delta_i)$, where $\vec{n}_i \in \mathbb{R}^2$ is the normalized direction vector toward the goal, $u_i \in \mathbb{N}$ quantifies the node utility (number of observable frontiers), and $\delta_i \in \{0, 1\}$ indicates the visitation status. The agent's observation is $o_t = (\mathbf{z}_C, G_t)$, where $\mathbf{z}_C = f_\theta(\mathcal{M}_t) \in \mathbb{R}^{N \times d}$ denotes contextual representations extracted by the pre-trained context encoder.

\subsubsection{Action Space} 
The action space $\mathcal{A}_t$ consists of selecting a navigable node from $V_t$. Upon action execution, the agent navigates to the selected waypoint and updates $\mathcal{M}_t$ and $G_t$.

\subsubsection{Policy Network} 
The policy network $\pi_\phi(a_t \mid o_t)$ integrates node-level features, historical trajectory information, and contextual spatial representations through a hierarchical attention mechanism. Each node's feature tuple is first embedded via linear projection to obtain node embeddings $\{\mathbf{e}_i^N\}_{i=1}^K$. Positional encoding is omitted from the embedding process since the nodes inherently contain spatial information. These embeddings are then processed through multi-head self-attention to capture inter-node relational features:
\begin{equation}
\mathbf{z}_N = \text{MHSA}(\mathbf{E}^N, \mathbf{E}^N, \mathbf{E}^N),
\end{equation}
where $\mathbf{E}^N = [\mathbf{e}_1^N; \mathbf{e}_2^N; \ldots; \mathbf{e}_K^N]$ denotes the stacked node embeddings.

To maintain temporal continuity, we employ an LSTM to recursively encode the agent's trajectory, producing a latent state representation $\mathbf{h}_t$. The hierarchical fusion proceeds as:
\begin{align}
\mathbf{h}_t^{(1)} &= \text{MHCA}(\mathbf{h}_t, \mathbf{z}_C, \mathbf{z}_C) + \mathbf{h}_t, \\
\mathbf{h}_t^{(2)} &= \text{MHCA}(\mathbf{h}_t^{(1)}, \mathbf{z}_N, \mathbf{z}_N) + \mathbf{h}_t^{(1)},
\end{align}
where $\text{MHCA}$ denotes multi-head cross-attention. This two-stage fusion enables the policy to first integrate global spatial context, then attend to task-relevant node features conditioned on this enriched representation.

Finally, we employ a pointer attention module~\cite{vinyals2015pointer} to generate action probabilities. The fused representation $\mathbf{h}_t^{(2)}$ serves as the query, while node embeddings $\mathbf{z}_N$ act as keys and values. The attention scores are normalized via softmax to produce a probability distribution over candidate nodes:
\begin{equation}
\pi_\phi(v_i \mid o_t) = \frac{\exp((\mathbf{h}_t^{(2)})^\top \mathbf{z}_i^N / \sqrt{d})}{\sum_{j=1}^K \exp((\mathbf{h}_t^{(2)})^\top \mathbf{z}_j^N / \sqrt{d})},
\end{equation}
where $d$ is the embedding dimension. The action is selected by sampling from this distribution during training or via argmax during inference.

\subsubsection{Reward} 
The reward balances task completion, trajectory efficiency, and goal-directed behavior:
\begin{equation}
r_t = r_{\text{goal}} \cdot \mathbb{1}_{\text{goal}} + \lambda_s r_s + \lambda_h r_h,
\end{equation}
where $r_{\text{goal}}$ is a fixed completion reward, $\mathbb{1}_{\text{goal}}$ is an indicator for goal arrival, $r_s = -1$ penalizes each time step to encourage efficiency, and $r_h = d(p_{t-1}) - d(p_t)$ provides a heuristic reward based on the change in geodesic distance to the goal. The hyperparameters $\lambda_s$ and $\lambda_h$ control the relative importance of these terms. An episode is considered successful when the agent reaches within 0.2m of the goal location, and terminates as a failure if the number of steps exceeds 128.

\subsubsection{Training} 
We train the navigation policy using Soft Actor-Critic (SAC)~\cite{christodoulou2019soft}, an off-policy actor-critic algorithm that maximizes entropy-regularized expected returns. The critic network shares the same architecture as the policy network up to the final layer, which outputs state-action value estimates. During training, we initialize the context encoder $f_\theta$ with weights pre-trained via the self-supervised framework (Section~\ref{sec:ssl}) and fine-tune it jointly with the policy network $\pi_\phi$.

\section{Experiment}
In this section, we validate the effectiveness and efficiency of our proposed MacroNav by answering the following critical questions:
\begin{itemize}
    \item Is our context encoder more effective than SotA pre-trained visual models for navigation?
    \item Does MacroNav demonstrate significant advantages over SotA navigation methods in real-world deployments?
    \item Are all three self-supervised tasks necessary for optimal performance?
\end{itemize}

\vspace{-5pt}
\subsection{Implementation Details}

\textbf{Self-Supervised Pre-training:} We train the context encoder on occupancy maps from HM3D~\cite{ramakrishnan2021habitat}, MP3D~\cite{chang2017matterport3d}, Gibson~\cite{xia2018gibson}, and HouseExpo~\cite{li2020houseexpo}, supplemented by synthetic maps generated programmatically. Since the generated maps follow a specific distribution, we carefully control the data mixing ratio to prevent overfitting during training.
We use embedding dimension $d=512$, $L=6$ encoder layers, $H=4$ attention heads, and patch size $P=8$. Training uses AdamW optimizer with learning rate $1\times10^{-4}$, batch size 256, and converges in $\approx$4 hours on an RTX-4090 GPU.

\textbf{RL Policy Training:} We employ the simulation environment from~\cite{liang2024hdplanner} and leverage Ray~\cite{moritz2018ray} for distributed training. Policy parameters: embedding dimension $d'=128$, 6 encoder layers, 8 attention heads, $K=20$ candidate nodes with $k=10$ for kNN, $r_{\text{local}}=3.0$m, $r_{\text{goal}}=20$, $\lambda_s=1$, $\lambda_h=2$. SAC uses learning rate $1 \times 10^{-5}$, discount factor $\gamma=0.99$, and target network update rate $\tau=0.005$. Training converges in $\approx$6 hours on an RTX-4090.

\subsection{Validations}

We first compare whether the context encoder pre-trained by the proposed SSL framework is more beneficial for navigation tasks than other visual models. To ensure fairness, we select baseline pre-trained models with approximately the same number of parameters. Notably, our context encoder is trained from scratch and does not utilize any pre-trained model as a starting point. The baselines include: the widely-used ResNet18/34~\cite{resnet}, ViT-small~\cite{dosovitskiyImageWorth16x162020} pre-trained on ImageNet (abbreviated as ViT-IN21K), DeiT-small~\cite{touvron2021training} (ViT-DeiT) and the SotA visual foundation model DINOv2-small~\cite{oquab2023dinov2} (ViT-DINO) that trained by self-supervised learning. 

We integrate each encoder with the same RL policy architecture and train on identical environments. Evaluation uses three difficulty levels (easy, medium, hard) in the simulator, measuring Success Rate (SR) and Success weighted by Path Length (SPL). Results are averaged over 200 episodes per difficulty level.

As shown in Table~\ref{tab:context_encoder_comparison_transposed}, our encoder achieves the best SR at all difficulty levels. While differences are modest in simple environments, performance gaps widen with complexity. In hard environments, our method outperforms the comparable-sized ViT-DINO by 9.5\% on SR and 11.7\% on SPL. This demonstrates significant improvements in complex scenarios. \textcolor{black}{To further verify that the learned representations capture general, algorithm-agnostic features, we also integrate our pre-trained encoder with TD3~\cite{fujimoto2018addressing} and PPO~\cite{schulman2017proximal} under identical policy architectures. The encoder consistently yields large gains across all three RL algorithms. SAC nevertheless achieves the best overall performance, likely due to its entropy-regularized objective promoting exploration and its off-policy nature enabling sample-efficient use of the rich spatial features.}

Notably, ViT-IN21K and ViT-DeiT perform worse than ResNet34 despite having comparable parameter counts, and even underperform training without the context encoder. We attribute this to the fact that when applied to occupancy maps, pure attention-based architectures lack the inductive biases inherent to CNNs for spatial pattern recognition. Without navigation-specific pre-training, these models struggle to learn effectively and instead introduce unfavorable noise. This observation is corroborated by attention visualizations (Figure~\ref{fig:attn_vit_comp}), which demonstrate that targeted navigation-specific pre-training is essential for ViT architectures to succeed in this domain.

\begin{figure*}[ht]
    \centering
    \resizebox{0.9\textwidth}{!}{ 
    \includegraphics{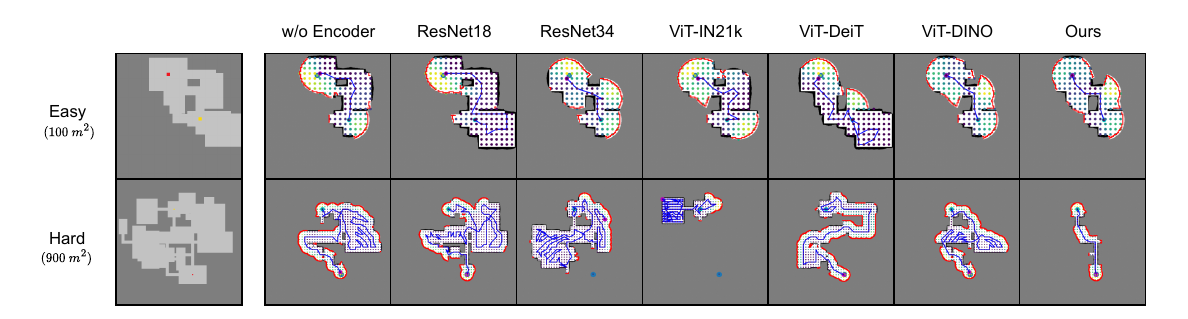}
    }
\caption{Trajectories of navigation policy with different context encoders in the unseen test environments.}
\label{fig:hard-path}  \vspace{-0.2cm}
\end{figure*}

Figure~\ref{fig:hard-path} presents a direct comparison of path planning strategies in the test environment. Our model demonstrates the ability to identify optimal paths and exhibits superior capability in recognizing critical shortcuts compared to other methods, indicating better environmental understanding than competing models.

\begin{table}[htbp]
\centering
\caption{
\textcolor{black}{
Performance comparison of different context encoders and representation transferability across RL algorithms. SR: Success Rate, SPL: Success weighted by Path Length. Unless otherwise noted, the policy is trained with SAC.
}
}
\label{tab:context_encoder_comparison_transposed}
\small
\setlength{\tabcolsep}{4pt}
\begin{tabular}{l|c|cc|cc|cc}
\toprule
\multirow{2}{*}{\textbf{Encoder}} & \multirow{2}{*}{\textbf{Params}} & \multicolumn{2}{c|}{\textbf{Easy}} & \multicolumn{2}{c|}{\textbf{Medium}} & \multicolumn{2}{c}{\textbf{Hard}} \\
\cmidrule(lr){3-8}
 & (M) & \textbf{SR} & \textbf{SPL} & \textbf{SR} & \textbf{SPL} & \textbf{SR} & \textbf{SPL} \\
\midrule
None & / & 90.9 & 77.1 & 78.8 & 63.7 & 49.7 & 32.9 \\
\midrule
ResNet18 & 11.7 & 81.8 & 56.5 & 67.7 & 33.9 & 35.7 & 14.0 \\
ResNet34 & 21.8 & \textbf{98.0} & 74.7 & 93.9 & 67.0 & 65.8 & 38.1 \\
ViT-IN21K & 22.1 & 65.7 & 45.2 & 53.5 & 24.5 & 32.2 & 12.8 \\
ViT-DeiT & 22.4 & 89.9 & 56.1 & 76.8 & 35.5 & 39.7 & 17.1 \\
ViT-DINO & 21.7 & \textbf{98.0} & \textbf{87.7} & 97.0 & 80.5 & 74.4 & 53.0 \\
\midrule
\rowcolor{gray!15}
\textbf{Ours} & 13.1 & \textbf{98.0} & 87.5 & \textbf{97.5} & \textbf{83.3} & \textbf{83.9} & \textbf{64.7} \\
\midrule
TD3 w/o enc. & / & 84.9 & 50.8 & 62.3 & 25.4 & 35.2 & 12.8 \\
\rowcolor{gray!15}
TD3 w/ Ours & 13.1 & \textbf{93.5} & \textbf{85.4} & \textbf{91.5} & \textbf{80.6} & \textbf{78.4} & \textbf{62.2} \\
PPO w/o enc. & / & 93.5 & 75.1 & 79.4 & 52.9 & 61.8 & 34.0 \\
\rowcolor{gray!15}
PPO w/ Ours & 13.1 & \textbf{98.0} & \textbf{80.3} & \textbf{94.0} & \textbf{76.2} & \textbf{82.4} & \textbf{58.3} \\
\bottomrule
\end{tabular}
\end{table}

\begin{figure}[ht]
    \centering
    \resizebox{0.48\textwidth}{!}{ 
    \includegraphics{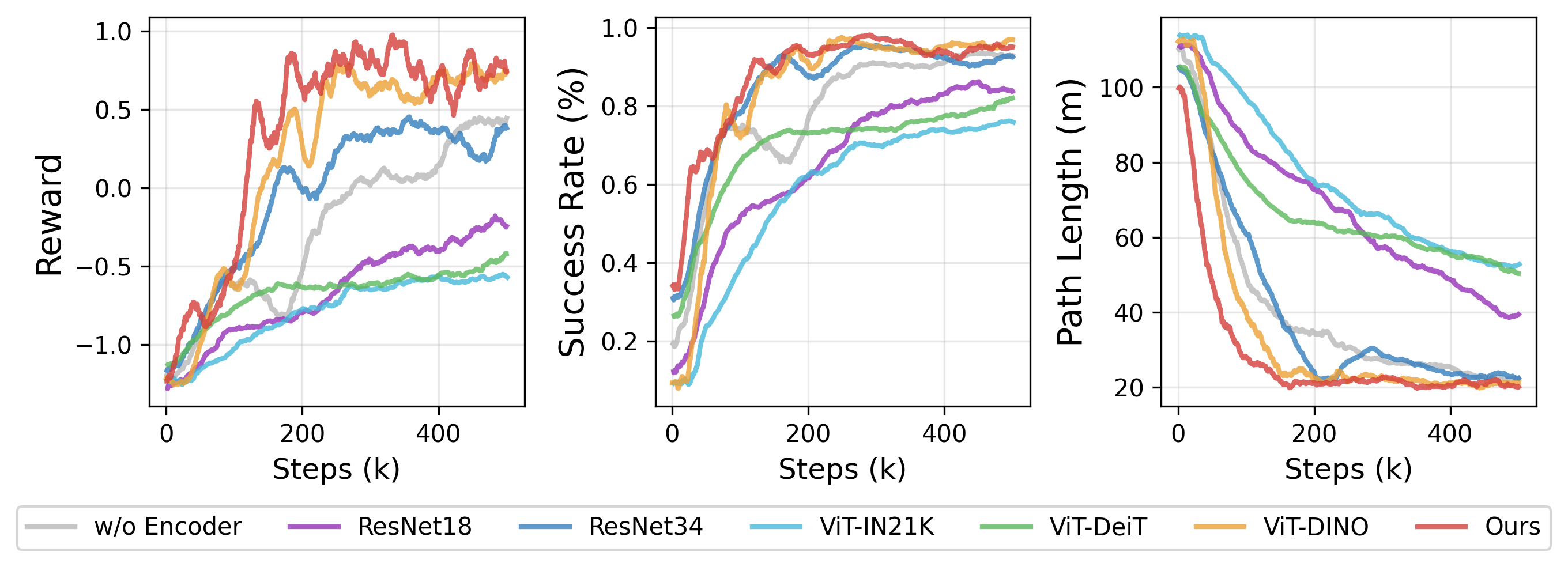}
    }
\caption{Comparison of training dynamics across different context encoders.}
\label{fig:training_dynamics} \vspace{-0.2cm}
\end{figure}

\begin{figure}[htp]
\centering \vspace{-0.2cm}
\includegraphics[width=0.48\textwidth]{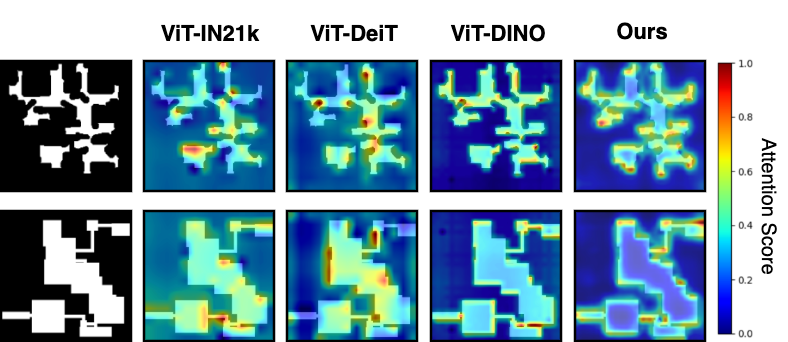}
\caption{\color{black} Attention visualization of different ViT-based context encoders.}
\label{fig:attn_vit_comp}
\end{figure}

Figure~\ref{fig:training_dynamics} shows that our method achieves the fastest training convergence, with particularly notable acceleration in path length reduction. This indicates that our representations better enable the RL policy to understand scenes and discover shortcuts.

To understand why our method improves navigation, we visualize attention patterns in Figure~\ref{fig:attn_vit_comp}. Remarkably, without explicit supervision, our encoder discovers important navigation features: it automatically segments effective navigation regions and attends strongly to key structures such as walls and corridors. ViT-DINO can also segment navigation areas but shows comparatively limited local geometric pattern recognition. This validates that our multi-task SSL framework successfully captures both global structure and local geometry.

 \vspace{-0.4cm}

\subsection{Real-World Experiments}

\begin{wrapfigure}{l}{0.26\textwidth}
    \vspace{-0.3cm}
    \centering
    \includegraphics[width=0.28\textwidth]{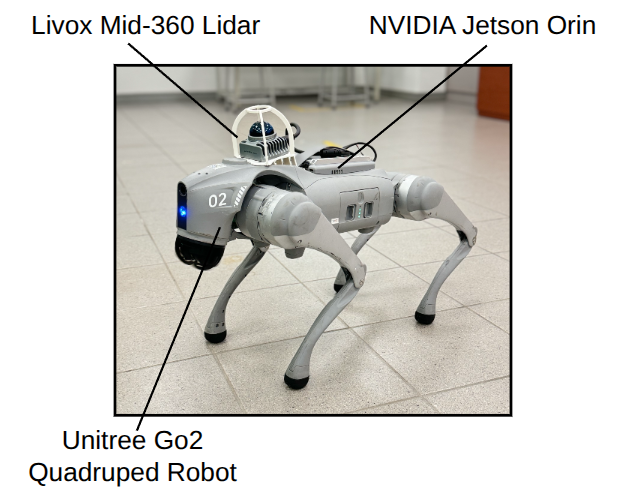}
    \caption{The robot platform.}
    \vspace{-0.2cm}
    \label{fig:robot_platform} \vspace{-0.3cm}
\end{wrapfigure}
\begin{figure}[htbp]
    \centering
    \includegraphics[width=0.5\textwidth]{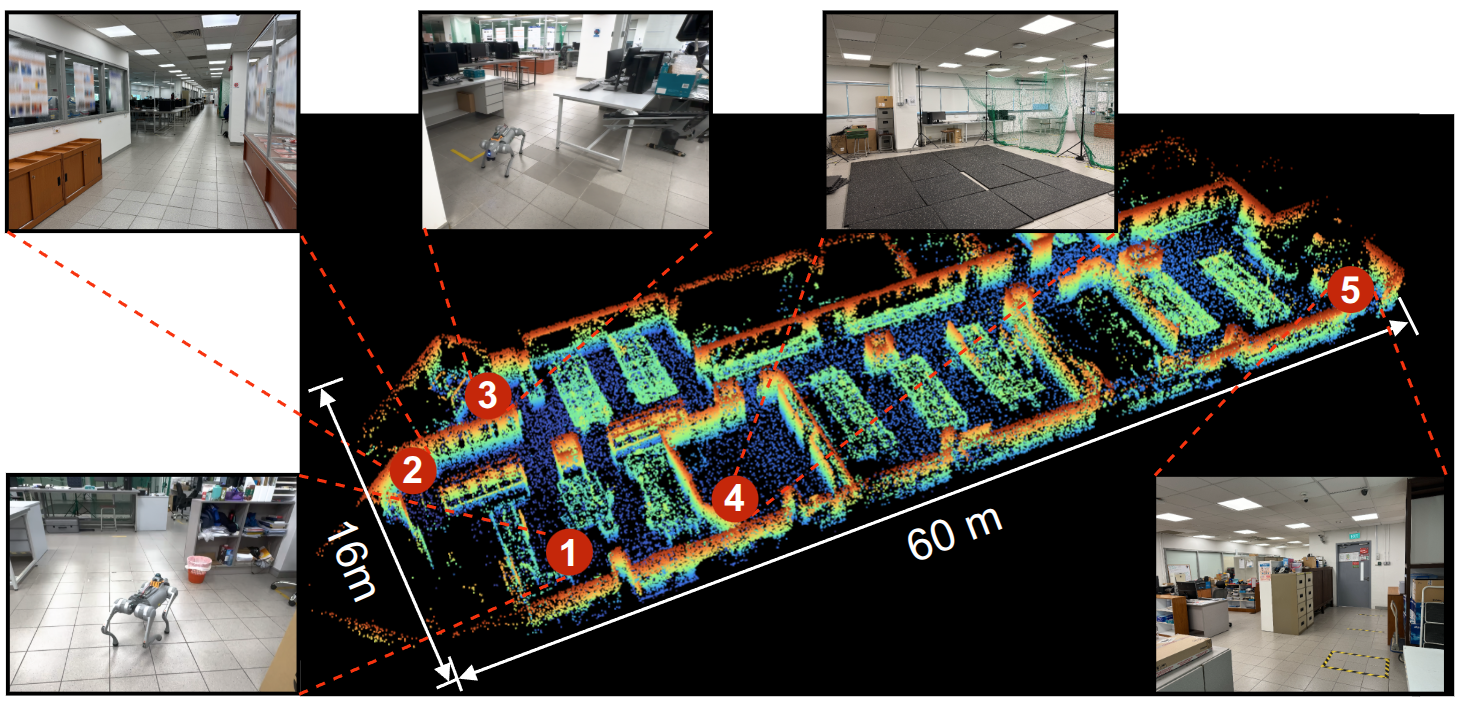}
    \caption{The real-world experiment environment.}
    \label{fig:real_env} \vspace{-0.2cm}
\end{figure}

\textbf{Experimental Setup:} We utilize a Unitree Go2 robot
as the experimental platform, equipped with a Livox Mid-360
LiDAR as the primary sensor, and employ FAST-LIO2~\cite{xu2022fast} for 
localization and mapping. All computations are executed on an 
NVIDIA Jetson Orin. 
\textcolor{black}{As shown in Figure~\ref{fig:real_env}, we define 5 key waypoints and construct 6 trajectory settings (start-goal pairs) covering small-scale to large-scale navigation for evaluating generalization. Each setting was tested twice, resulting in 12 
trials in total. Success is defined as reaching the goal within 0.2m; 
failure occurs after 128 decision steps without reaching the goal.}

\textbf{Baselines:} \textbf{FAR Planner}~\cite{yang_far_2022}, a popular non-learning-based method using visibility graph and graph search, proven to be effective in real-world scenarios; \textbf{NavRL}~\cite{xu2025navrl}, a SotA end-to-end RL navigation method; and \textbf{HD Planner}~\cite{liang2024hdplanner}, a SotA RL-based navigation method for unknown environments. All methods use the official open-source code. For our method and non-end-to-end baselines (FAR, HD Planner), we use DWB~\cite{nav2} as the local planner with identical parameters for fair comparison.

\textbf{Evaluation Metrics:} SR, SPL, single-step planning time, and computational resource utilization (CPU, GPU, RAM). Computational efficiency is critical for deployment alongside other modules (SLAM, object detection).

\textcolor{black}{
Table~\ref{tab:real_exp_comp} presents real-world results. MacroNav achieves 91.7\% SR and 73.4\% SPL, representing absolute improvements of 8.4 percentage points (pp) in SR and 54.8 pp in SPL over HD Planner, and 16.7 pp in SR and 9.7 pp in SPL over FAR Planner. While FAR Planner has lower planning time, our method maintains significant advantages over learning-based approaches with 0.036s planning time while satisfying real-time requirements.
}

MacroNav achieves the lowest CPU and GPU utilization, leaving ample resources for other modules. In contrast, FAR Planner's high CPU usage (84.0\%) stems from real-time visibility graph construction. In the 6th trajectory (large-scale scene), FAR's extensive computations caused SLAM divergence due to insufficient resources. Our approach demonstrates superior scalability through local topological graphs and dense spatial representations.

\begin{figure*}[ht]
\centering
\includegraphics[width=\textwidth]{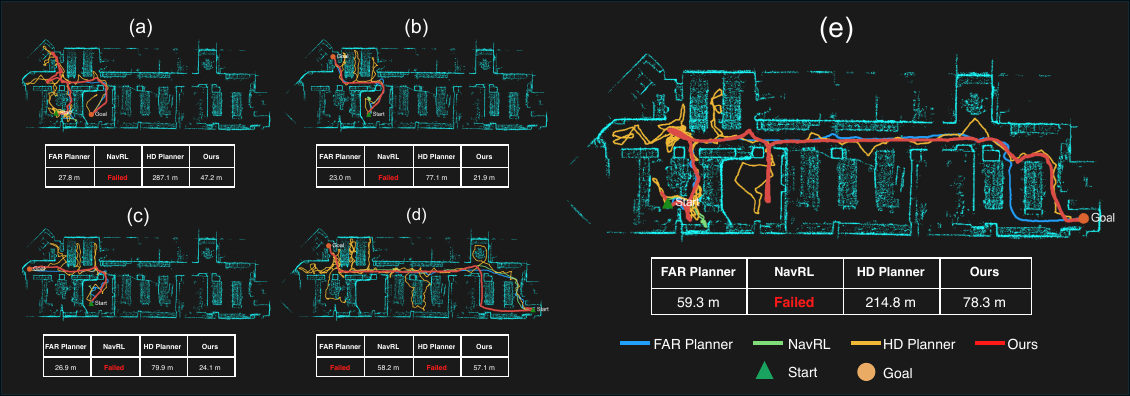}
\caption{\color{black} Comparison of representative trajectories of different methods in the real-world experiments.}
\label{fig:real_exp_traj} \vspace{-0.2cm}
\end{figure*}

Figure~\ref{fig:real_exp_traj} visualizes trajectories in real-world. Our method exhibits significantly shorter paths with minimal detours and accurately identifies shortcuts. HD Planner achieves high SR but requires substantial detours in nearly every episode, indicating inadequate spatial understanding. This demonstrates the significant advantage of our multi-scale environmental representation.

Surprisingly, NavRL achieves only 16.7\% SR in our scenarios. Failure mode analysis reveals NavRL frequently becomes trapped in local optima: despite successfully avoiding obstacles, the policy fails to guide the agent out of dead ends, leading to conflicting behaviors between obstacle avoidance and global routing. We attribute this to the lack of explicit global planning and insufficient environmental understanding, validating the necessity of our context-aware framework.

\textcolor{black}{
We further analyze MacroNav's failure cases to identify remaining limitations. The dominant observed failure occurs during long-range navigation, where the policy exhibits oscillatory waypoint selection and exceeds the step limit, which we attribute to the lack of trajectory-level optimization in the current RL formulation. Beyond this, real-world layouts beyond the pretraining coverage may yield biased context representations, and the sim-to-real gap of the simulator may induce out-of-distribution states that further degrade decision-making.
}

\begin{table*}[ht]
\centering
\caption{Comparison of planning methods across different performance metrics.}
\label{tab:real_exp_comp}
\begin{tabular}{lcccccc}
\toprule
\textbf{Method} & \textbf{SR (\%)} $\uparrow$ & \textbf{SPL (\%)} $\uparrow$ & \textbf{$T_{plan}$ (s)} $\downarrow$ & \textbf{CPU (\%)} $\downarrow$ & \textbf{GPU (\%)} $\downarrow$ & \textbf{RAM (\%)} $\downarrow$ \\
\midrule
FAR Planner & 75.0 & 63.7 & \textbf{0.004} $\pm$ \textbf{0.003} & 84.0 $\pm$ 7.8 & N/A & \textbf{39.3} $\pm$ {2.1} \\
NavRL & 16.7 & 14.5 & 0.089 $\pm$ 0.034 & 80.5 $\pm$ 8.3 & 22.3 $\pm$ 17.3 & 67.3 $\pm$ \textbf{1.8} \\
HD Planner & 83.3 & 18.6 & 0.261 $\pm$ 0.289 & 67.3 $\pm$ 8.3 & 2.2 $\pm$ 5.4 & 54.8 $\pm$ 5.2 \\
\midrule
\textbf{Ours} & \textbf{91.7} & \textbf{73.4} & {0.036 $\pm$ 0.020} & \textbf{48.9 $\pm$ 5.5} & \textbf{1.3 $\pm$ 2.6} & {46.1 $\pm$ 5.7} \\
\bottomrule
\end{tabular}
\end{table*}

\subsection{Ablation Studies}
We conduct ablation experiments on our multi-task self-supervised learning framework to verify the necessity of each task combination. We pre-train the context encoder using single tasks and pairwise combinations, then jointly train them with the same RL policy and evaluate navigation performance in identical test environments (hard difficulty).

\begin{table}[htbp]
\centering
\caption{Ablation study on training tasks for prediction and navigation.}
\label{tab:ssl_abla}
\setlength{\tabcolsep}{3.5pt} 
\resizebox{\columnwidth}{!}{ 
\begin{tabular}{ccc|ccc|cc}
\toprule
\multicolumn{3}{c|}{\textbf{Tasks}} & \multicolumn{3}{c|}{\textbf{Prediction}} & \multicolumn{2}{c}{\textbf{Navigation}} \\
\textbf{FOV} & \textbf{SPM} & \textbf{MAE} & \textbf{FOV$\downarrow$} & \textbf{SPM$\downarrow$} & \textbf{MAE$\downarrow$} & \textbf{SR$\uparrow$} & \textbf{SPL$\uparrow$} \\
\midrule
-          & -          & -          & -              & -              & -              & 38.2           & 17.2           \\
\checkmark &            &            & 0.317          & 0.449          & 2.601          & 79.9           & 57.3           \\
           & \checkmark &            & 0.336          & \textbf{0.118} & 2.467          & 70.4           & 51.0           \\
           &            & \checkmark & 1.237          & 1.277          & 1.530          & 72.9           & 47.9           \\
\checkmark & \checkmark &            & 0.313          & 0.135          & 2.757          & 72.3           & 52.1           \\
\checkmark &            & \checkmark & 0.333          & 0.251          & 0.961          & 81.9           & 62.4           \\
           & \checkmark & \checkmark & 0.338          & 0.120          & 0.954          & 76.4           & 61.1           \\
\midrule
\checkmark & \checkmark & \checkmark & \textbf{0.293} & 0.123          & \textbf{0.867} & \textbf{83.9}  & \textbf{64.7}  \\
\bottomrule
\end{tabular}
}
\end{table}

\textbf{Reconstruction Accuracy:} As shown in Table~\ref{tab:ssl_abla}, combining all three tasks yields the highest reconstruction accuracy for FOV and MAE, with only SPM showing slight degradation. This indicates the three tasks are mutually related and complementary at the representation level.

\textbf{Navigation Performance:} The navigation results in Table~\ref{tab:ssl_abla} further validate this complementarity. Using all three tasks achieves the highest SR and SPL. Among individual tasks, FOV performs best, likely because local geometric understanding is most directly relevant to waypoint selection. The FOV+MAE combination improves performance, whereas FOV+SPM degrades it. We attribute this to inherent tension between global scene understanding and local geometric recognition. However, MAE alleviates this tension by enforcing understanding of visible regions, bridging the gap between global and local representations.

\textcolor{black}{
\section{Conclusion}
We present MacroNav, a novel framework that combines a context encoder trained by multi-task self-supervised representation learning and an RL-based navigation policy for efficient autonomous navigation in unknown environments. Extensive experiments demonstrate significant improvements over state-of-the-art methods.
\textcolor{black}{
Several directions remain for future work. First, the current representation is purely geometric and built from single-frame occupancy maps, omitting both temporal cues useful for anticipating dynamic obstacles and semantic cues required for tasks such as object-goal navigation; extending the SSL framework to temporal sequences and semantic maps is a promising direction. 
Second, as revealed by our failure analysis, the RL policy can exhibit oscillatory behavior in long-range navigation due to the absence of trajectory-level optimization, which we plan to address through reward shaping that encourages trajectory smoothness and sequential consistency.
}
}

\FloatBarrier

\bibliographystyle{IEEEtran}
\bibliography{reference}



 



\end{document}